# A Locally Adaptive Interpretable Regression

**Lkhagvadorj Munkhdalai[1], Tsendsuren Munkhdalai[2] and Keun Ho Ryu[3*]**
[1]Database and Bioinformatics Laboratory, School of Electrical and Computer Engineering, Chungbuk National University, Cheongju, 28644, Republic of Korea
[2]Microsoft Research, Montreal, QC H3A 3H3, Canada
[3]Faculty of Information Technology, Ton Duc Thang University, Ho Chi Minh City, 70000, Vietnam
lhagii@dblab.chungbuk.ac.kr, tsendsuren.munkhdalai@microsoft.com, khryu@tdtu.edu.vn

## Abstract

The main shortcoming of linear regression is the weak predictive performance due to the linearity. However, the linearity leads to interpretable models, which are easy to quantify and describe. In this work, we introduce a locally adaptive interpretable regression (LoAIR). In LoAIR, a meta model parameterized by neural networks predicts percentile of a Gaussian distribution for the regression coefficients for a rapid adaptation. Our experimental results on public benchmark datasets show that our model not only achieves comparable or better predictive performance than the other state-of-the-art baselines but also discovers some interesting relationships between input and target variables such as a parabolic relationship between $CO_2$ emissions and Gross National Product (GNP). Therefore, LoAIR is a step towards bridging the gap between econometrics, statistics and machine learning by improving the predictive ability of linear regression without depreciating its interpretability.

## 1 Introduction

A linear regression identifies a linear relationship between the target and input variables. This linearity makes the estimation procedure simple, and most importantly, easy to understand the interpretation of the correlation between variables [Goldberger, 1962; Andrews, 1974]. On the other hand, statistical properties of the Ordinary Least Squares (OLS) estimator (unbiasedness, efficiency, consistency and asymptotic normality) make it trustworthy. These superiorities and statistical guarantees led the linear regression to use in many fields including economics, biology, management, chemical science, and social science. However, the major drawback of linear regression is linearity as well. The linear relationships are hardly restricted and usually oversimplify how complex reality is; therefore, the predictive ability of linear regression is often not good. In contrast, the predictive capacity of deep neural networks is usually high, but due to its nonlinear black box nature, these models are difficult to interpret [LeCun, Bengio and Hinton, 2015; Ribeiro, Singh and Guestrin, 2016].

In this work, we propose a locally adaptive interpretable regression (LoAIR) model to achieve both high predictive accuracy and interpretability. We apply deep neural networks as a meta-learner to predict percentile of Gaussian distribution for the regression coefficients to make them rapidly adaptable. The overall architecture of LoAIR is shown in figure 1. LoAIR consists of two main components: we first perform the OLS to obtain regression coefficients and their standard errors. Second, we apply deep neural networks to predict the probabilities for finding the Gaussian critical value to adapt each regression coefficient. Based on the predicted probabilities and the standard error of regression coefficients, we rebuild the regression equation by using adapted coefficients. In the LoAIR framework, the estimated coefficients are unbiased as well as adapted within their confidence intervals. This helps avoid the model overfitting and keep the model interpretable. At same time, the predictive ability of the model is improved.

We extensively studied the predictive performance and the model interpretability on several benchmark datasets for the regression task. Our proposed model not only achieves comparable or better predictive performance than the other state-of-the-art baselines but also reveals some interesting relationships between input and target variables such as a parabolic relationship between $CO_2$ emissions and Gross National Product (GNP).

The rest of the paper is organized with discussion of related work in Section 2, the proposal of the LoAIR in Section 3, and experiments in Section 4.

## 2 Related Work

Attempts to develop locally adaptive regression models have begun much earlier [Cleveland 1979; Cleveland and Devlin, 1988; Trevor and Tibshirani, 1996; Dirk and Hastie, 2000]. Those local regression models can be categorized into three types – nearest neighbor, weighted average, and locally weighted regressions [Atkeson, Moore and Stefan, 1997]. Nearest neighbor, local models mostly use *k* closest points for a query point to estimate its underlying function for the output value [Trevor and Tibshirani, 1996; Emmanuel 2000;

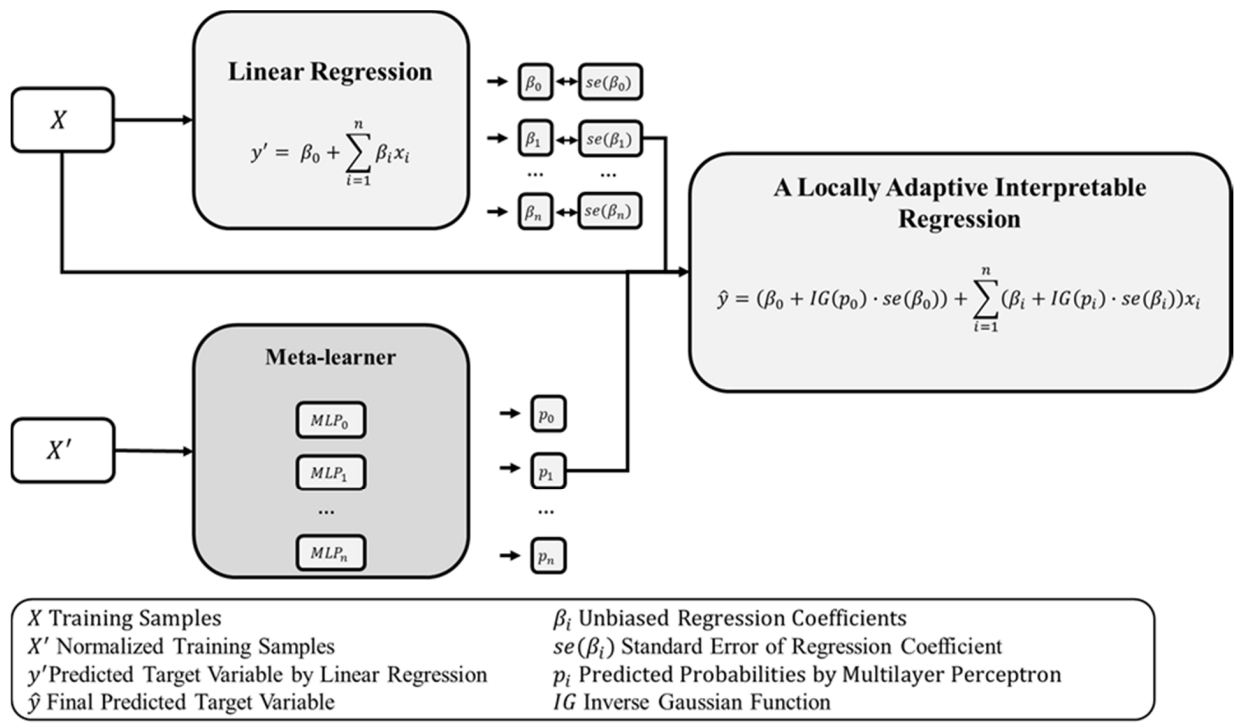

Figure 1: Overall architecture of the LoAIR model.

Biau, Cérou and Guyader 2010; Zhao and Lifeng, 2019; Chen and Ioannis, 2019]. Weighted average local models take a weighted average output of nearby points, weighted by the inverse of their distance to the query point [Nadaraya, 1964; Fan, 1992; Nguyen *et al.,* 2009; Moon, Nathan and Sung-Eui, 2014]. Locally weighted regression (LWR) is fitted on a local set of points using a distance-weighted regression [Cleveland 1979; Cleveland and Devlin, 1988]. The similarity with the LoAIR to LWR is that this model does not learn a fixed set of parameters. However, those local adaptive regression models are similar with memory-based learning where they require keeping the entire training set to predict unknown values. Thus, these models are computationally intensive for large datasets. Instead, we design a meta-learner based on deep neural networks to adapt the regression coefficients rapidly and our model could be more efficient on large datasets. Utilizing one neural network to produce parameters for another neural network has been studied earlier in *meta-learning* field [Hinton and Plaut, 1987; Schmidhuber, 1993 and 1992; Paul, 2001].

From the meta-learning perspective, we train meta model to explain its underlying base model (linear regression) parameters. Munkhdalai and Hong [2017] recently proposed Meta Networks (MetaNet) that learns to fast parameterize underlying neural networks for rapid generalizations. Our method is based on the idea of the MetaNet that uses fast-weights, which has successfully been used in the meta-learning context for rapid adaptation [Munkhdalai and Adam 2018; Munkhdalai *et al.,* 2018; Munkhdalai *et al.,* 2019; Ha, Andrew and Le, 2017; Munkhdalai *et al.,* 2019]. Our meta-learner estimates fast probabilities for finding the Gaussian critical value for each regression coefficient in order not to undermine the model interpretability as well as to overestimate the model.

Another related line of work focuses on Bayesian regression. Inference in the Bayesian linear model is based on the posterior distribution over regression coefficients, computed by Bayes' rule [Neal, 2012; Faul and Tipping 2002; Raftery, Madigan and Hoeting 1997]. In other words, a classical treatment of the linear regression problem seeks a point estimate of regression coefficient. By contrast, the Bayesian approach characterizes the uncertainty in regression coefficients through a probability distribution. However, while we use point estimation, we also have a range of possible values for the regression coefficients as called confidence interval. Our model adapts the regression coefficients within their confidence intervals. A range for the regression coefficients named a credible interval in Bayesian inference can be used in our framework as well.

Finally, to the best of our knowledge, our work is the first attempt to improve the predictive ability of linear regression by adapting deep neural networks.

## 3 A Locally Adaptive Interpretable Regression

Our LoAIR consists of two main phases – linear regression and adaptation with meta-learner (see figure 1). We first perform simple linear regression on training set to obtain unbiased regression coefficients and their standard errors. Second, we train deep neural networks as a meta-learner on nor-malized training set to predict the probability for finding percentile of Gaussian distribution for each regression coefficient. Finally, we reconstruct the linear regression equation by using the adapted regression coefficients.

## 3.1 Linear Regression

Given a set of dataset $(x_1, y_1), ... (x_n, y_n)$ of $n$ observations, a linear regression model estimates the $\beta$ coefficients that provide the best linear fits between the dependent variable ($y_i$) and $p$ independent variables ($x_{i1}, ... x_{ip}$). The model for linear regression is:

$$y_i = \beta_0 + \sum_{j=1}^{p} \beta_j x_{ij} + \varepsilon_i \quad (1)$$

where $\varepsilon_i$ are independent, identically distributed ($i.i.d.$) random variables with $\mathbb{E}\{\varepsilon_i\} = 0, \mathbb{E}\{\varepsilon_i^2\} = \sigma^2$ and bounded third moment.

The regression coefficients can simply be computed by using the OLS estimator:

$$\hat{\beta}_{OLS} = (X_n^\top X_n)^{-1} X_n^\top y_n \quad (2)$$

where $X_n = [x_1^\top, ... x_n^\top] \in \mathbb{R}^{n \times p}$ is the design matrix and $y_n = [y_1, ... y_n] \in \mathbb{R}^n$. The regression coefficients estimated from data are subject to sampling uncertainty. In other words, the true value of the regression coefficient can never be estimated from the sample data [Wonnacott and Wonnacott, 1990; Chen, 1993]. Instead, we would construct confidence interval for each regression coefficient:

$$CI_{\alpha/2}^{\beta_j} = [\hat{\beta}_j - IG_{\alpha/2} * se(\hat{\beta}_j), \hat{\beta}_j + IG_{\alpha/2} * se(\hat{\beta}_j)] \quad (3)$$

where $\alpha$ is the significance level, $IG$ is the inverse Gaussian distribution and $se(\hat{\beta}_j)$ is the standard error of the regression coefficient $\hat{\beta}_j$.

Before we introduce the LoAIR model, we conduct a simulation study based on confidence intervals to realize a better understanding of our idea. We investigate the relationship between $CO_2$ emission and gross national product (GNP). The data between 1990 and 2015 is chosen as a training set and data in 2016 is a test set. We then perform linear regression on training set and evaluate the error by using three countries' data randomly sampled from test set. In addition, we generate the regression coefficients 5000 times within their confidence intervals using Monte Carlo simulation to calculate the errors for the selected three samples. Figure 2 shows the error surface for the simulation study. We can now clearly see that better predictions can be done by adapting the linear regression coefficients in the range between their confidence intervals. In order to perform this adaptation process, we must be able to predict the appropriate significance level (probability) for both each regression coefficient and each observation. Therefore, we propose a novel deep neural network architecture for finding the appropriate significance level or probability for each regression coefficient to make it adaptable.

## 3.1 Meta-learner for LoAIR

We use a simple Multilayer Perceptron (MLP) neural network as a meta-learner [Munkhdalai and Hong 2017; Munkhdalai and Adam 2018; Munkhdalai *et al.*, 2018; Munkhdalai *et al.*, 2019]. Input of MLP can be normalized $p$ independent variables ($x'_{i1}, ... x'_{ip}$) and output should be the predicted probability ($prob_i$). Since we predict probability for finding the critical value of Gaussian distribution, the activation function of output layer can be sigmoid ($\sigma$). Thus:

$$prob = \sigma(\omega \cdot h(x'; \theta) + b) \quad (4)$$

where $x'$ is normalized input, $prob$ denotes the predicted probability and $\omega, \theta$, and $b$ denote the weight parameters of MLP and $h(*)$ is the output of hidden layers.

Note that the output of the sigmoid function can be either 0 or 1, in which case the inverse Gaussian distribution will be undefined. We then make additional smoothing on the output of sigmoid.

$$prob = \frac{\sigma(\omega \cdot h(x'; \theta) + b) + \epsilon}{1 + \tau} \quad (5)$$

where $\epsilon, \tau$ ($\tau > \epsilon$) are smoothing parameters and these parameters should be close to 0. We can also set upper and lower confidence intervals for the regression coefficients by adjusting these smoothing parameters.

Recall that we pick the estimated regression coefficients and their standard errors as input after performing linear regression. So we can easily reconstruct the original regression equation during the learning process of the neural networks:

$$\hat{y} = (\hat{\beta}_0 + IG(prob_0) \cdot se(\hat{\beta}_0)) + \sum_{i=1}^{p}(\hat{\beta}_i + IG(prob_i) \cdot se(\hat{\beta}_i))x_i \quad (6)$$

where $x_i$ is *i-th* independent variable (not normalized). From here, we can easily design our loss function as follows:

$$\mathcal{L}(\omega, \theta, b) = loss_{MSE}(m(x, x', \hat{\beta}, se(\hat{\beta}); \omega, \theta, b), y) \quad (7)$$

where $loss_{MSE}$ is the mean is squared error (MSE) and $m(*)$ is the LoAIR with parameters $\omega, \theta$, and $b$.

In addition, our meta-learner model can consist of one or multiple MLPs, and the output of meta-learner should be equal to the number of independent variables. Both architectures can easily be trained with stochastic gradient descent (SGD) optimization.

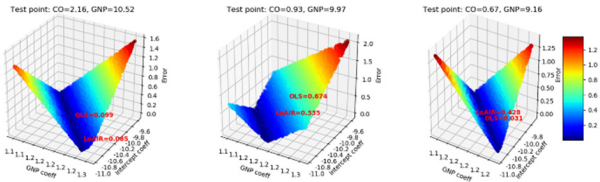

Figure 2: The error surface for the simulation study.

## 4 Experimental Results

In this section, we use public benchmark datasets (see Table 1) to compare the predictive performance of LoAIR to the other state-of-the-art baselines. We also apply LoAIR to a real-world data to demonstrate the model interpretability.

| Dataset | #Observations | #Variables |
|---|---|---|
| Boston Housing | 506 | 13 |
| Concrete Strength | 1030 | 8 |
| Energy Efficiency | 768 | 8 |
| Kin8nm | 8192 | 8 |
| Naval Propulsion | 11,934 | 16 |
| Power Plant | 9568 | 4 |
| Protein Structure | 45,730 | 9 |
| Wine Quality | 1599 | 11 |
| Yacht Hydrodynamics | 308 | 6 |
| Year Prediction MSD | 515,345 | 90 |

Table 1: Summary of datasets

### 4.1 Predictive Performance

We first evaluate the predictive accuracy of LoAIR and compare it to deep learning and regression baselines. We utilize the same datasets in [Hernández-Lobato and Adams, 2015] and [Yarin and Ghahramani, 2016] to directly compare our results with theirs as shown in Table 2. Hernández-Lobato and Adams [2015] proposed a scalable method for learning Bayesian neural networks, called probabilistic backpropagation (PBP) and compared with variational inference (VI) method in Bayesian neural networks [Graves, 2011] as well as standard stochastic gradient descent via back-propagation (BP). Lastly, Yarin and Ghahramani [2016] replicated the experiment set-up in [Hernández-Lobato and Adams, 2015] and compared their proposed theoretical framework casting dropout training in deep neural networks as approximate Bayesian inference (Dropout) to PBP and VI. In this work, we adopt the results of these studies as deep learning baselines. For regression baselines, locally weighted scatterplot smoother (Loess) [Cleveland and Devlin, 1988], Bayesian regressions (Bayesian) [Raftery, Madigan and Hoeting 1997] and OLS are used for the performance comparison.

In our LoAIR, we need to define deep neural network architecture and other hyperparameters for meta-learner. We trained two types of architectures - separated MLPs for each regression coefficient (Multiple MLPs) and only one MLP with multiple outputs that equal to the number of the regression coefficients (shared MLP). Meta-learner consists of three hidden layers with {64, 64, and 16} neurons for each MLP, respectively. For hyper-parameters, we set the learning rate to 0.001 and the maximum epoch number for training to 5000. In addition, an Early Stopping algorithm was used for finding the optimal epoch number based on given other hyperparameters. The smoothing parameters were chosen as $\epsilon$=1e-06 and $\tau$=1e-05. We configured the same model settings for all datasets and datasets were partitioned into three parts; i.e., training (75%), validation (15%) and test sets (10%). All experiments were averaged on five random splits of the data (apart from Year Prediction MSD for which one split was used).

Our proposed LoAIR model outperformed deep learning baselines on 5 out of 10 datasets and showed comparable performance on the other datasets. Regression baseline models underperformed our model on most of the datasets. We significantly improved the predictive accuracy of linear regression. Typically, the predictive accuracy of OLS is weaker than that of the Loess, but after the adaptation that we did, its predictive ability encourages dramatically. The aim of this experiment is to demonstrate how the predictive ability of linear regression improves after the adaptation and we can now observe it. The next part of the experiments will show the interpretability of the LoAIR model.

### 4.2 Model Interpretability

In this section, we consider a real-world dataset, which is the link between $CO_2$ emission and gross national product ($GNP$) dataset [Douglas and Selden, 1995] that we examined in Section 3.1. The source of data is the official web page of Our World in Data. As we mentioned before, the data between 1990 and 2015 is the training and data in 2016 is the test set. To investigate the link between $CO_2$ emission and $GNP$, we estimate two different regression equations as follows:

$$CO_2 = \beta_0 + \beta_1 * GNP \quad (8)$$
$$CO_2 = \beta_0 + \beta_1 * GNP + \beta_1 * GNP^2 \quad (9)$$

Generally, assuming that there are positive linear and negative parabolic relationships between $CO_2$ emission and $GNP$.

| Dataset | Deep learning baselines | | | | Our LoAIR | | Regression baselines | | |
|---|---|---|---|---|---|---|---|---|---|
| | BP | VI | PBP | Dropout | Multiple MLPs | Shared MLP | OLS | Loess | Bayesian |
| Boston Housing | 3.23±0.19 | 4.32±0.29 | 3.01±0.18 | **2.97±0.85** | 3.11±0.50 | 3.40±0.84 | 4.68±1.09 | 3.68±0.53 | 4.84±1.11 |
| Concrete Strength | 5.98±0.22 | 7.19±0.12 | 5.67±0.09 | 5.23±0.53 | **5.04±1.06** | 5.92±1.17 | 10.7±0.83 | 7.18±0.62 | 10.6±0.85 |
| Energy Efficiency | 1.1±0.07 | 2.65±0.08 | 1.80±0.05 | 1.66±0.19 | **0.47±0.04** | 0.53±0.06 | 3.22±0.28 | 4.24±0.29 | 3.21±0.28 |
| Kin8nm | **0.09±0.00** | 0.10±0.00 | 0.10±0.00 | 0.10±0.00 | 0.14±0.00 | 0.16±0.00 | 0.20±0.00 | 0.13±0.00 | 0.20±0.00 |
| Naval Propulsion | **0.001±0.00** | 0.01±0.00 | 0.01±0.00 | 0.01±0.00 | 0.005±0.00 | 0.005±0.00 | 0.005±0.00 | 0.014±0.00 | 0.005±0.00 |
| Power Plant | 4.18±0.04 | 4.33±0.04 | 4.12±0.03 | 4.02±0.18 | **3.80±0.17** | 3.82±0.17 | 4.49±0.12 | 4.10±0.11 | 4.49±0.12 |
| Protein Structure | 4.54±0.03 | 4.84±0.03 | 4.73±0.01 | 4.36±0.04 | **4.26±0.05** | 4.45±0.08 | 5.17±0.02 | NA | 5.17±0.02 |
| Wine Quality | 0.64±0.01 | 0.65±0.01 | 0.64±0.01 | **0.62±0.04** | 0.67±0.02 | 0.67±0.02 | 0.67±0.03 | 0.64±0.03 | 0.67±0.03 |
| Yacht Hydrodynamics | 1.18±0.16 | 6.89±0.67 | 1.02±0.05 | 1.11±0.38 | 1.02±0.43 | **0.84±0.16** | 8.50±0.83 | 8.86±1.08 | 8.47±0.83 |
| Year Prediction MSD | 8.93±NA | 9.034±NA | 8.879±NA | **8.849±NA** | 13.00±NA | 14.47±NA | 17.7±0.17 | NA | 17.7±NA |

Table 2: Average test performance in RMSE.

| Variables | Linear term | Linear and quadratic terms |
|---|---|---|
| Intercept | -10.76*** | -18.81*** |
| log(GNP) | 1.27*** | 3.13*** |
| log(GNP)^2 |  | -0.105*** |
|  |  |  |
| R-squared | 0.839 | 0.85 |
| Prob (F-statistic) | 0 | 0 |

Table 3: Estimated coefficients of OLS regression.

| Model | RMSE | MAE | R-squared |
|---|---|---|---|
| LoAIR with linear term | 0.591 | 0.482 | 0.851 |
| OLS with linear term | 0.607 | 0.483 | 0.843 |
| LoAIR with linear and quadratic terms | 0.593 | 0.481 | 0.85 |
| OLS with linear and quadratic terms | 0.598 | 0.488 | 0.848 |

Table 4: The prediction performance on $CO_2$ emission test dataset.

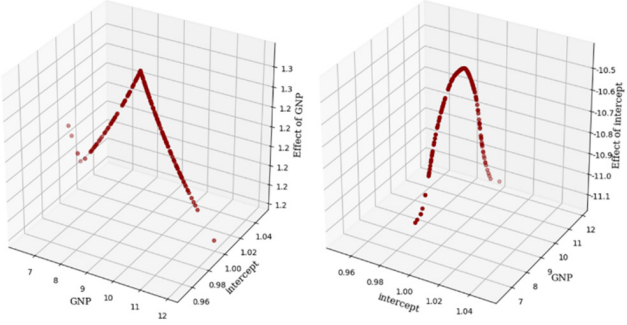

Figure 3: The relationship between *GNP*, intercept and the estimated coefficients by LoAIR model.

Theoretically, the Environmental Kuznets Curve (EKC) hypothesis postulates an inverted-U-shaped relationship between $CO_2$ emission and *GNP* [Douglas and Selden, 1995]. Our estimates of Eqs.(8) and (9) are reported in Table 3. The regression coefficients are consistent with EKC hypothesis. We then trained LoAIR model on these two OLS results and reported the prediction performance in Table 4.

Our LoAIR model showed slightly better performance than both two OLS results. Finally, we capture the relationship between $CO_2$ emission and *GNP* from the LoAIR model. Figure 3 shows the effect of *GNP* (left) and intercept (right) on $CO_2$ emission for Eq (8). We can easily see that *GNP* and intercept are parabolic with $CO_2$ emission. When *GNP* goes up to 9.16, it intensively increases $CO_2$ emission, then when *GNP* is higher than 9.16 its effect on $CO_2$ emission starts to decrease. For intercept, average $CO_2$ emission increases up to a certain level as *GNP* goes up; after that, it decreases. In Eq (9), we added the quadratic term of *GNP* as an independent variable, and the prediction performance of OLS improves. Although the prediction performance of the LoAIR model has not changed much, its interpretability is shifted as shown in figure 4. We can now see that the parabolic relationship between $CO_2$ and *GNP* on Eq (8) has transformed to linear. Our model can also measure how much $CO_2$ will change due to the change in *GNP* for each observation. Therefore, LoAIR promoted the interpretability of OLS as well.

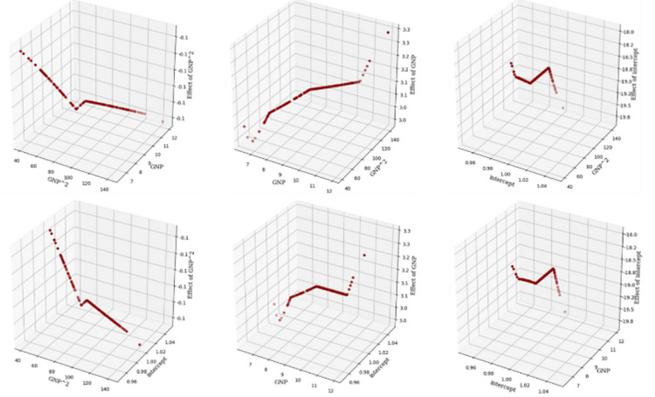

Figure 4: The relationship between *GNP*, the quadratic term of *GNP*, intercept and the estimated coefficients by LoAIR model.

## 5 Conclusion

In this work, we introduced a novel locally adaptive interpretable regression called LoAIR. LoAIR augments a linear regression model with a meta-level deep neural network that predicts percentile of Gaussian distribution for each regression coefficient for rapid adaptation. We conducted an extensive set of experiments to show the interpretability and predictive power of LoAIR. Our model significantly improved the predictive power of OLS and highlighted interesting relationships between input and output variables.

A more general AI-based solution to the interpretably issue is to train another model to learn to explain the main predictive model. As LoAIR is a first attempt along this line of research, we believe that it opens an exciting venue for future work.

## Acknowledgments

This research was supported by the Basic Science Research Program through the National Research Foundation of Korea (NRF) funded by the Ministry of Science, ICT & Future Planning (No.2017R1A2B4010826 and 2019K2A9A2A06020672).